\documentclass{article}
\usepackage{spconf,amsmath,graphicx,booktabs, multirow,makecell}
\usepackage{comment}
\begin{document}

\title{Improving Contextual Coherence in Variational Personalized and Empathetic Dialogue Agents}

\name{Jing Yang Lee$^{\dagger}$ \qquad Kong Aik Lee$^{\ddagger}$ \qquad Woon Seng Gan$^{\dagger}$}
  
  \address{$^{\dagger}$ School of Electrical and Electronic Engineering, Nanyang Technological University, Singapore \\
      $^{\ddagger}$ Institute for Infocomm Research, A*STAR, Singapore}


\maketitle

\begin{abstract}
In recent years, latent variable models, such as the Conditional Variational Auto Encoder (CVAE), have been applied to both personalized and empathetic dialogue generation. Prior work have largely focused on generating diverse dialogue responses that exhibit persona consistency and empathy. However, when it comes to the contextual coherence of the generated responses, there is still room for improvement. Hence, to improve the contextual coherence, we propose a novel Uncertainty Aware CVAE (UA-CVAE) framework. The UA-CVAE framework involves approximating and incorporating the aleatoric uncertainty during response generation. We apply our framework to both personalized and empathetic dialogue generation. Empirical results show that our framework significantly improves the contextual coherence of the generated response. Additionally, we introduce a novel automatic metric for measuring contextual coherence, which was found to correlate positively with human judgement.
\end{abstract}


\section{Introduction}
The ultimate goal of open-domain dialogue agents is to engage humans in seamless, natural conversation. To achieve this goal, dialogue agents are expected to generate responses which demonstrate empathy and persona consistency. Hence, the tasks of personalized \cite{lee-etal-2021-generating, su19b_interspeech, liu-etal-2020-impress} and empathetic (or emotional) dialogue generation \cite{song-etal-2019-generating,li-etal-2020-empdg, zhou-wang-2018-mojitalk, huang-etal-2018-automatic} has received significant research attention in recent years. Traditionally, neural personalized or empathetic open-domain dialogue agents demonstrate a tendency to generate generic and repetitive responses. To address the lack of diversity in the generated responses, numerous approaches have explored the application of latent variable models, particularly the Conditional Variational Auto Encoder (CVAE), to personalized and empathetic dialogue generation. 
Empirical results from prior works \cite{DBLP:journals/corr/abs-2111-11363, 9406340, Song2019ExploitingPI,wu-etal-2020-guiding,zhao-etal-2017-learning,zhou-wang-2018-mojitalk} have shown that CVAE-based models have largely succeeded in generating diverse dialogue responses. 
Despite the gains in response diversity, there is still much room for improvement, in particular, when it comes to contextual coherence.

To improve the contextual coherence of the responses generated by CVAE-based agents, we propose approximating and incorporating uncertainty during the generation process. There are two types of uncertainty: aleatoric and epistemic uncertainty \cite{NIPS2017_2650d608}. While aleatoric uncertainty captures the uncertainty in the input data, epistemic uncertainty quantifies the uncertainty due to the model. Typically, aleatoric uncertainty is predicted as a model output, and the epistemic uncertainty is obtained via ensemble methods \cite{NIPS2017_9ef2ed4b} or monte carlo simulations \cite{pmlr-v48-gal16}. Aleatoric uncertainty can be further categorized as either homoscedastic or heteroscedastic. Homoscedastic, or task-dependant, uncertainty refers to input invariant noise which differs from task to task. On the other hand, heteroscedastic, or data-dependant, uncertainty quantifies the uncertainty (or noise) inherent in each input and is thus unique to each input. 

Hence, we introduce an Uncertainty-Aware CVAE (UA-CVAE) framework, which involves approximating and incorporating the heteroscedastic aleatoric uncertainty via an uncertainty aware latent variable. 
We also propose an Utterance Entailment (UE) score that quantifies the contextual coherence of an agent's responses. Experimental results show that our framework significantly improves overall response coherence. The remainder of this paper is organized as follows: An overview of aleatoric uncertainty approximation in CVAE-based dialogue agents is provided in Section 2. In Section 3, the UA-CVAE framework and UE score are described in detail. Experimental results and the conclusion are provided in sections 4 and 5 respectively.

\section{Uncertainty in Dialogue Generation}
For our discussion, let $X = \{x_{0,0},...,x_{M-1,N-1}\}$ refers to the dialogue context ($M$ refers to the number of utterances in the context, and $N$ denotes the number tokens in a given utterance), and $Y = \{y_{0},...,y_{L-1}\}$ refers to the generated dialogue response ($L$ represents the number of tokens). Similarly, $\bar{Y}$ will refer to the reference dialogue response. Additionally, we use $z$ and $c$ to denote the sampled latent variable and any external information (persona description or emotion label) respectively.
\subsection{CVAE-based Dialogue Generation}
In a CVAE-based dialogue agent, the generated response $Y$ is conditioned on the the dialogue context $X$, the external information $c$ (persona description or emotion label), as well as latent variable $z$, which is randomly sampled from latent Gaussian distribution $p(z|X, c)$. An overview is provided in Fig 1. The stochasticity introduced by $z$ improves the diversity of the generated dialogue.
\begin{figure}[]
    \centering
    \includegraphics[width=0.8\columnwidth, height=20mm]{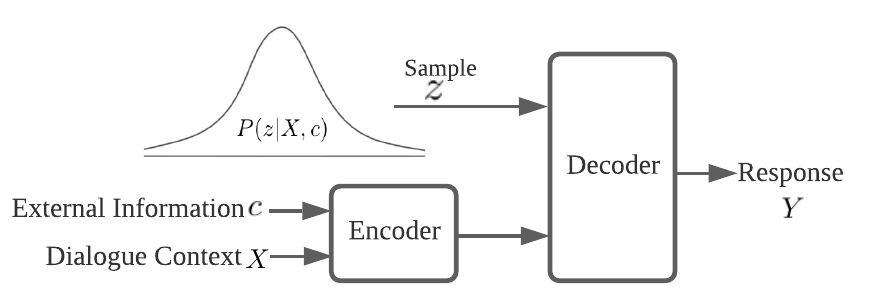}
    \caption{Overview of a CVAE-based dialogue agent.}
    \label{fig:UA-CVAE}
\end{figure}
\subsection{Aleatoric Uncertainty Approximation}
 In this section, we show that for a CVAE-based dialogue agent, the variance $\sigma^{2}$ of the latent Gaussian distribution is an approximation of the \emph{aleatoric uncertainty} of the generated dialogue response. For a CVAE-based dialogue agent, the latent variable $z$ is assumed to follow an isotropic multivariate Gaussian distribution:
\begin{equation}
    z \sim \mathcal{N}(\mu,\sigma^{2}\textbf{I})
\end{equation}
where $\mu$ and $\sigma^{2}$ refer to the mean and variance vectors of the latent distribution. The CVAE-based dialogue generation process can be represented by the following conditional distribution:
\begin{equation}
    p(Y,z|X, c) = p(Y|z,X, c)p(z|X, c)
\end{equation}
In a CVAE-based dialogue agent, both $p(Y|z,X, c)$ and $p(z|X, c)$ will be defined by neural networks parameterized by $\pi$ and $\theta$ respectively. In particular, as $z$ is a Gaussian random variable and $p(z|X, c) = \mathcal{N}(\mu,\sigma^{2}\textbf{I})$, the network $p_{\theta}(z|X, c)$ will be tasked with generating a mean $\mu$ and a variance $\sigma^{2}$. 
Hence, since the variance $\sigma^{2}$ predicted by the neural network $p_{\theta}(z|X, c)$ is dependant on the input data, $\sigma^{2}$ would capture the uncertainty inherent in the input $X$ and $c$ i.e., the heteroscedastic aleatoric or data-dependant uncertainty. This is similar to the uncertainty estimation approach introduced in \cite{pmlr-v48-gal16}, where a data dependant function of the uncertainty is learned. From (2), we observe that the response generation process $p(Y,z|X, c)$ and $p(z|X, c)$ are both conditioned only on $X$ and $c$. Hence, we conclude that the aleatoric uncertainty of the generated dialogue $p(Y,z|X, c)$ can be approximated by the variance $\sigma^{2}$, which captures the aleatoric uncertainty of $p(z|X, c)$. 
\begin{figure}[ht!]
\centering
    \includegraphics[width=0.8\columnwidth, height=60mm]{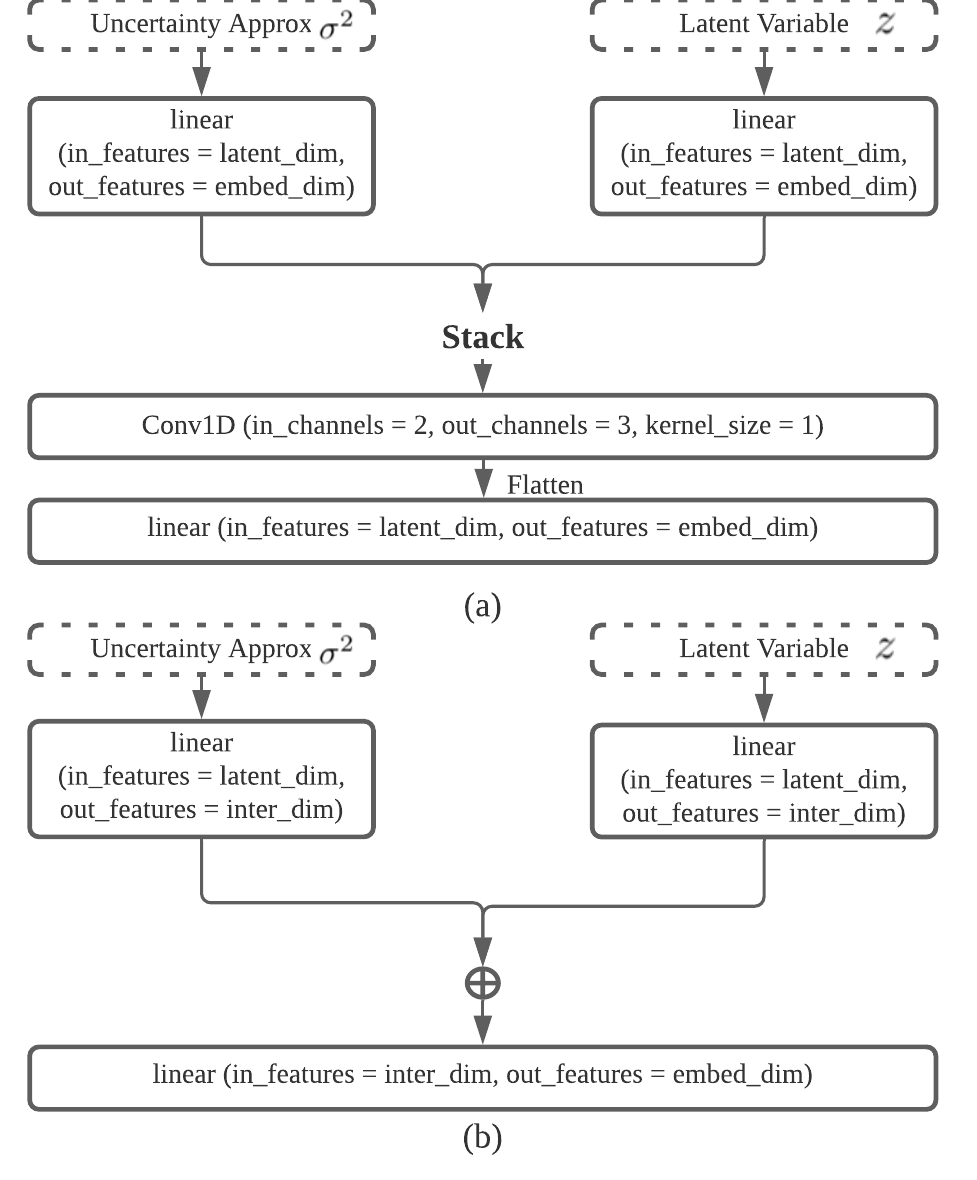}
    \caption{(a) Overview of the combination network in UA-CVAE(C). (b) Overview of the combination network in UA-CVAE(M). $\oplus$ indicates element-wise addition. Both networks output $z_{\text{u}}$. The embedding dimension (embed\_dim), intermediate dimension (inter\_dim) and the latent dimension (latent\_dim) are fixed at 768, 384 and 256 respectively.}
    \label{fig:Combination Network}
\end{figure}
\section{Uncertainty-Aware CVAE}

\begin{figure*}[ht!]
    \centering
    \includegraphics[width=14cm,height=3.5cm]{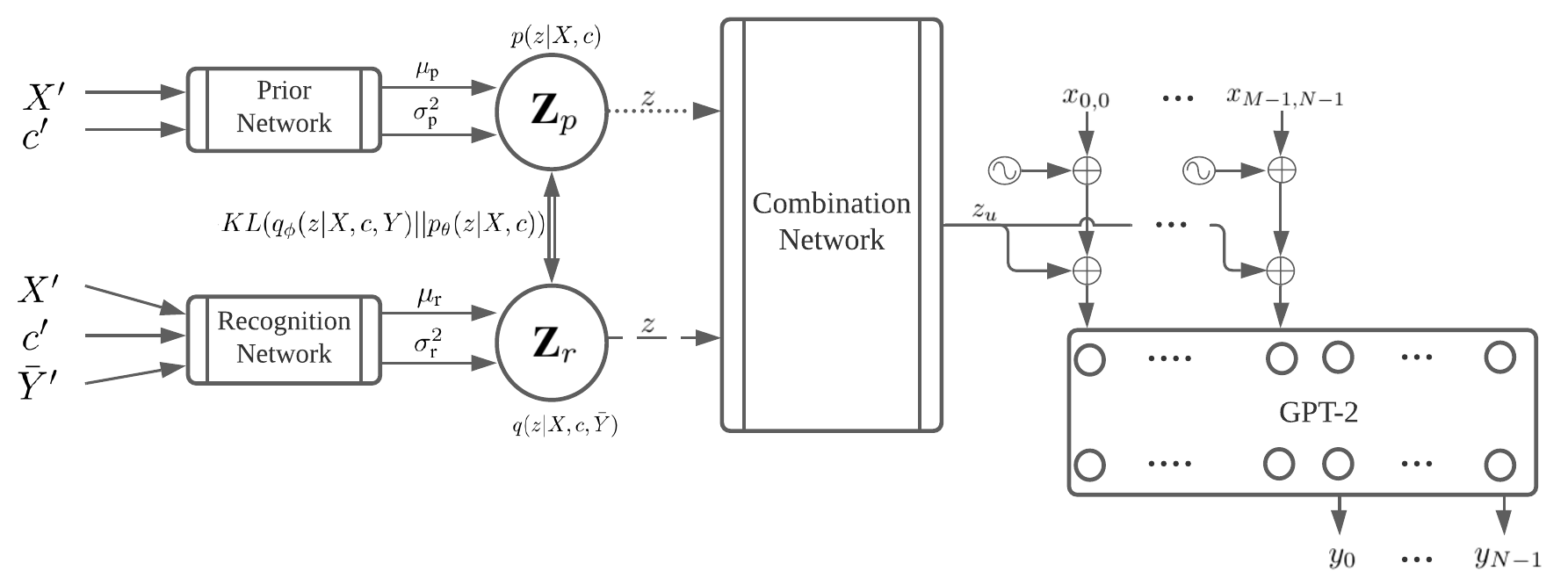}
    \caption{Overview of the UA-CVAE framework. The dashed and dotted connection occurs in only the training and inference stages respectively.}
    \label{fig:UA-CVAE}
\end{figure*}
As mentioned, the conditional distributions $p(Y|z,X, c)$ and $p(z|X, c)$ are defined by neural networks parametrized by $\pi$ and $\theta$ respectively. A GPT-2 pretrained model \cite{radford2019language} is used to obtain the sequence embeddings of $X$, $c$ and $\bar{Y}$, denoted by $X'$, $c'$ and $\bar{Y}'$ respectively. The prior network $p_{\theta}(z|X, c)$ is composed of a single-layer MLP tasked with generating the mean $\mu_{\text{p}}$ and the log-variance $log(\sigma^{2}_{\text{p}})$ of the latent Gaussian distribution:
\begin{equation}
\begin{bmatrix} \mu_{\text{p}} \\log(\sigma^{2}_{\text{p}}) \end{bmatrix} = W_{\text{p}} \begin{bmatrix} c' \\ X'\end{bmatrix} + b_{\text{p}}
\end{equation}
where $W_{\text{p}}$ and $b_{\text{p}}$ refer to the weights and bias of the prior network. Before being fed into the prior network, the sequence embedding of the dialogue context and the external information are obtained via a GPT-2 pretrained model. The variance, is attained by computing the exponential of the log-variance $log(\sigma^{2}_{\text{p}})$. To obtain $z$, the latent Gaussian distribution is randomly sampled via the reparametrization trick to ensure differentiability. During training, the true posterior distribution $q(z|X,c,\bar{Y})$ is approximated by another MLP parameterized by $\phi$. This network, termed the recognition network $q_{\phi}(z|X,c,\bar{Y})$, also consists of a single layer MLP: 
\begin{equation}
\begin{bmatrix} \mu_{\text{r}} \\log(\sigma^{2}_{\text{r}}) \end{bmatrix} = W_{\text{r}} \begin{bmatrix} c' \\ X'\\ \bar{Y}'\end{bmatrix} + b_{\text{r}}
\end{equation}
where $W_{\text{r}}$ and $b_{\text{r}}$ refer to the weights and bias of the recognition network. Similarly, Before being fed into the prior network, the sequence embedding of the dialogue context and the external information are obtained via a GPT-2 pretrained model \cite{radford2019language}. The KL divergence between the Gaussian distributions generated by the prior and recognition network will be incorporated in the final loss function. 
Then, the latent variable $z$ and the variance vector $\sigma^{2}_{\text{p}}$ (inference) or $\sigma^{2}_{\text{r}}$(training), an approximation of the \emph{aleatoric uncertainty} of the generated dialogue response, is fed to the combination network which outputs the uncertainty aware latent variable $z_{\text{u}}$. By conditioning the decoder on the uncertainty aware latent variable $z_{\text{u}}$, the decoder would possess an awareness of the uncertainty or noise inherent in $X$ and $c$. This allows the decoder to recognize and adapt to uncertain or noisy dialogue contexts and persona description/emotion labels, reducing the number of contextually incoherent responses. We implement 2 variants of the combination network (Fig 2), resulting in 2 variants of the UA-CVAE framework: UA-CVAE(M) and UA-CVAE(C).  While the combination network in UA-CVAE(M) constitutes a several linear layers, the combination network in UA-CVAE(C) involves a 1D convolution layer. Then, $z_{\text{u}}$ is added to the positional encoding and token-level embedding at each decoding step, and fed to the response decoder $p_{\pi}(Y|z,X, c)$ which constitutes a GPT-2 pretrained language model \cite{radford2019language}.

Since training a CVAE would require an intractable marginalization over $z$, Stochastic Gradient Variational Bayes (SGVB) is used to train the UA-CVAE. Additionally, due to the vanishing latent variable problem, we include the Bag-of-Words (BoW) loss in the loss function \cite{zhao-etal-2017-learning}. The final loss function is as follows:
\begin{equation}
\begin{split}
L(X,&Y,c,z, z_{\text{u}}) = - KL(q_{\phi}(z|X,c,Y)||p_{\theta}(z|X,c)) \\ 
&+E_{q_{\phi}(z|X,c,Y)}[log\:p(Y|X,c,z_{\text{u}})]\\ 
&+E_{q_{\phi}(z|X,c,Y)}[log\:p(Y_{bow}|X,c,z_{\text{u}})] 
\end{split}
\end{equation}
where $Y_{bow}$ refers to the generated dialogue response BoW. A overview is provided in Fig 3.

\subsection{Utterance Entailment Score}
Prior work have shown that the task of Natural Language Inference, which involves evaluating if a given premise entails or contradicts a predefined hypothesis, correlates with human judgement with regard to contextual coherence. Our metric, termed the Utterance Entailment (UE) score,  requires finetuning a BERT model on the SNLI corpus and applying the resultant model to the dialogue context and generated response. Specifically, given a single utterance in the dialogue context $x_{i}$ and a generated response $Y$, the model judges if the premise $x_{i}$ and the hypothesis $Y$ are contradictory, entailing or neutral. Then, a rating of 1,-1  or 0 is assigned when the premise and hypothesis are entailing, contradictory or neutral respectively. For each response, the final Utterance Entailment score is computed by summing all assigned ratings. Hence, for each dialogue example, the UE score can be computed as follows:
\begin{equation}
UE(X,Y) = \sum_{i = 1}^{M}BERT_{NLI}(x_{i},Y)
\end{equation}
where $X$ and $Y$ refer to the dialogue context and generated response respectively. The final UE-score for a given test set is obtained by averaging the UE-scores of each dialogue example.

\section{Experiments \& Results}
\begin{table*}
\centering
\caption{Results of automatic evaluation.}
\scalebox{0.65}{
\begin{tabular}{cccccccccccc} 
\toprule
                                                                                         & Model      & PPL               & Avg Length        & Rouge-l (F1)      & Rouge-l (R)       & Rouge-l (P)       & METEOR           & Distinct-1       & Distinct-2       & Distinct-3       & UE-score          \\ 
\hline
\multirow{7}{*}{\rotcell{ConvAI2}}                                                       & seq2seq    & 42.5615           & 10.3481           & 0.0115           & 0.0672           & 0.0064           & 0.3109           & 0.3927           & 0.6870           & 0.7509           & 0.0002            \\
                                                                                         & HRED       & 33.0022           & 10.3594           & 0.0126           & 0.0732           & 0.0069           & 0.3023 & 0.4134           & 0.7531           & 0.8204           & 0.0006            \\
                                                                                         & VHRED      & 39.6422           & 10.5012           & 0.0110           & 0.0636           & 0.0061           & \textbf{0.3154}           & 0.4339           & 0.8142           & 0.8817           & 0.0008   \\
                                                                                         & VHCR       & 32.1812           & 10.8626           & 0.0114           & 0.0660           & 0.0063           & 0.3099           & 0.4364           & 0.7641           & 0.8183           & 0.0004            \\
                                                                                         & GPT-2      & 15.5766           & 9.5820            & 0.1434           & \textbf{0.1846 } & 0.1236           & 0.2294           & 0.4496           & 0.8160           & 0.8705           & 0.0540            \\
                                                                                         & CVAE       & 14.2457           & 11.0069           & 0.1172           & 0.1406           & 0.1073           & 0.2631           & 0.4773           & \textbf{0.8729 } & \textbf{0.9245 } & 0.0406            \\
                                                                                         & UA-CVAE(M) & 14.5423           & \textbf{11.7421 } & \textbf{0.1291 } & 0.1517           & 0.1200           & 0.2576           & 0.4562           & 0.8554           & 0.9149           & 0.0648            \\
                                                                                         & UA-CVAE(C) & \textbf{14.1253}  & 11.3658           & 0.1169           & 0.1475           & \textbf{0.1245 } & 0.2507           & \textbf{0.4981 } & 0.8648           & 0.9106           & \textbf{0.0906 }  \\ 
\hline
\multirow{7}{*}{\rotcell{\begin{tabular}[c]{@{}c@{}}Empathetic\\Dialogues\end{tabular}}} & seq2seq    & 52.9023           & 10.9731           & 0.0054           & 0.0355           & 0.0030           & 0.3243           & 0.4468           & 0.7561           & 0.8130           & 0.0103            \\
                                                                                         & HRED       & 68.2843           & 11.0345           & 0.0053           & 0.0353           & 0.0029           & \textbf{0.3311 } & 0.4444           & 0.8333           & 0.8988           & 0.0095            \\
                                                                                         & VHRED      & 55.4275           & 11.0139           & 0.0045           & 0.0297           & 0.0025           & 0.3305           & 0.4396           & 0.8149           & 0.8798           & 0.0089            \\
                                                                                         & VHCR       & 63.4752           & 11.2758           & 0.0034           & 0.0416           & 0.0034           & 0.3017           & 0.4884           & 0.7668           & 0.8112           & 0.0090            \\
                                                                                         & GPT-2      & 16.6456           & 7.3657            & 0.0390           & 0.0676           & 0.0299           & 0.2525           & 0.4848           & 0.7902           & 0.8315           & 0.1081            \\
                                                                                         & CVAE       & 16.3177           & 8.0514            & 0.0745           & 0.1087           & 0.0651           & 0.2333           & 0.5139           & 0.8715           & 0.9328           & 0.1074            \\
                                                                                         & UA-CVAE(M) & \textbf{15.2021 } & 9.8180            & \textbf{0.0867 } & \textbf{0.1132 } & 0.0817           & 0.2913           & \textbf{0.5526 } & \textbf{0.9149 } & \textbf{0.9370 } & 0.1198            \\
                                                                                         & UA-CVAE(C) & 17.1932           & \textbf{12.1111 } & 0.0859           & 0.1045           & \textbf{0.0825 } & 0.2794           & 0.4922           & 0.8877           & 0.9239           & \textbf{0.1257 }  \\
\bottomrule
\end{tabular}}
\end{table*}

\begin{table}
\centering
\caption{Results of human evaluation. Kappa scores generally range from 0.5 to 0.6, indicating moderate agreement.}
\scalebox{0.585}{
\begin{tabular}{lccccccccc} 
\toprule
\multirow{2}{*}{}                                                                                            & \multirow{2}{*}{Model} & \multicolumn{4}{c}{UA-CVAE(M)}                 & \multicolumn{4}{c}{UA-CVAE(C)}                  \\ 
\cline{3-10}
                                                                                                             &                        & Win  & Tie  & Loss & \multicolumn{1}{l}{Kappa} & Win  & Tie  & Loss & \multicolumn{1}{l}{Kappa}  \\ 
\hline
\multirow{8}{*}{\rotcell{ConvAI2}}                                                                           & seq2seq                & 53\% & 45\% & 2\%  & 0.54                      & 55\% & 39\% & 6\%  & 0.59                       \\
                                                                                                             & HRED                   & 49\% & 42\% & 9\%  & 0.49                      & 58\% & 38\% & 4\%  & 0.46                       \\
                                                                                                             & VHRED                  & 43\% & 41\% & 16\% & 0.52                      & 43\% & 51\% & 6\%  & 0.50                       \\
                                                                                                             & VHCR                   & 42\% & 38\% & 20\% & 0.64                      & 51\% & 41\% & 8\%  & 0.59                       \\
                                                                                                             & GPT-2                  & 37\% & 58\% & 5\%  & 0.57                      & 41\% & 45\% & 14\% & 0.51                       \\
                                                                                                             & CVAE                   & 34\% & 49\% & 17\% & 0.43                      & 44\% & 45\% & 11\% & 0.47                       \\
                                                                                                             & UA-CVAE(M)             & -    & -    & -    & -                         & 29\% & 48\% & 23\% & 0.63                       \\
                                                                                                             & UA-CVAE(C)             & 23\% & 48\% & 29\% & 0.63                      & -    & -    & -    & -                          \\ 
\hline
\multicolumn{1}{c}{\multirow{8}{*}{\rotcell{\begin{tabular}[c]{@{}c@{}}Empathetic\\Dialogues\end{tabular}}}} & seq2seq                & 50\% & 42\% & 8\%  & 0.42                      & 58\% & 32\% & 10\% & 0.66                       \\
\multicolumn{1}{c}{}                                                                                         & HRED                   & 53\% & 43\% & 4\%  & 0.49                      & 64\% & 29\% & 7\%  & 0.58                       \\
\multicolumn{1}{c}{}                                                                                         & VHRED                  & 57\% & 38\% & 5\%  & 0.53                      & 57\% & 38\% & 5\%  & 0.55                       \\
\multicolumn{1}{c}{}                                                                                         & VHCR                   & 59\% & 33\% & 8\%  & 0.54                      & 61\% & 37\% & 2\%  & 0.60                       \\
\multicolumn{1}{c}{}                                                                                         & GPT-2                  & 33\% & 52\% & 15\% & 0.64                      & 52\% & 29\% & 19\% & 0.49                       \\
\multicolumn{1}{c}{}                                                                                         & CVAE                   & 39\% & 47\% & 14\% & 0.56                      & 34\% & 53\% & 13\% & 0.54                       \\
\multicolumn{1}{c}{}                                                                                         & UA-CVAE(M)             & -    & -    & -    & -                         & 31\% & 56\% & 13\% & 0.64                       \\
\multicolumn{1}{c}{}                                                                                         & UA-CVAE(C)             & 13\% & 56\% & 31\% & 0.64                      & -    & -    & -    & -                          \\
\bottomrule
\end{tabular}}
\end{table}
\subsection{Corpora}
We benchmark our framework on the ConvAI2 \cite{dinan2019second} and EMPATHETICDIALOGS \cite{rashkin2019towards} corpora. For EMPATHETICDIALOGS, the human speaker describes a personal situation and the dialogue agent is expected to generate an empathetic response. Each dialogue example is provided an emotion label (e.g., `excited'). ConvAI2 consists of dialogues where two interlocutors are trying to get to know each other. Each interlocutor is assigned a persona description consisting of multiple persona statements (e.g., `I love dogs', `I like eggs'). 
\subsection{Implementation \& Baselines}
In our implementation of UA-CVAE(M) and UA-CVAE(C), the response decoder consists of a GPT-2 pretrained language model \cite{radford2019language} which is finetuned during training. We utilize the GPT-2 base model from HuggingFace consisting of 12 layers, 12 attention heads as well as a hidden dimension of 768. The size of the latent variable $z$ and $z_{\text{u}}$ was fixed at 256. The Adam optimizer (learning rate = 0.0001, batch size = 16) was used. The maximum utterance length and maximum number of turns were fixed at 40 and 4 respectively.

In addition to \textbf{UA-CVAE(C)} and \textbf{UA-CVAE(M)}, we implement the following baselines: \textbf{Seq2seq} with attention applied to the GRU decoder, Hierarchical Recurrent Encoder-Decoder (\textbf{HRED}) \cite{10.5555/3016387.3016435}, Variational VHRED (\textbf{VHRED}) \cite{Serban_Sordoni_Lowe_Charlin_Pineau_Courville_Bengio_2017}, Variational
Hierarchical Conversation RNN (\textbf{VHCR}) \cite{park-etal-2018-hierarchical}, \textbf{GPT-2}, and the \textbf{CVAE} with a GPT-2 decoder.
\section{Results \& Discussion}
\subsection{Results}
We evaluate the generated dialogue with the following automatic metrics (Table 1): PPL, ROUGE \cite{lin-2004-rouge}, METEOR \cite{banerjee-lavie-2005-meteor}, intra-response Distinct-n (1,2,3) \cite{li-etal-2016-diversity}, and  the UE-Score. The average response length is also provided. 
For human evaluation (Table 2), we engage 4 graduates to evaluate the contextual coherence of the generated responses. Specifically, each graduate was told to compare a response generated by a UA-CVAE variant with a response generated by the baselines, and select the more contextually coherent response.
\subsection{Discussion}

Based on the human evaluation results and the automatic metrics, we can conclude that there is a positive correlation between the proposed UE-score and human judgement. UA-CVAE(M)/(C) generally achieve a lower percentage of wins and higher percentage of losses against baselines with higher UE-scores such as GPT-2 and CVAE.  

It is also apparent that UA-CVAE(M) and UA-CVAE(C) outperformed all implemented baselines in terms of contextual coherence. Furthermore, the Distinct-n, Rouge and METEOR scores achieved by UA-CVAE(M) and UA-CVAE(C) were comparable to that of the CVAE baseline, suggesting comparable response diversity and fluency. In terms of UE-score, compared to the CVAE baseline, UA-CVAE(M) and UA-CVAE(C) achieved a 60\% and 121\% improvement respectively on ConvAI2. However, on EMPATHETICDIALOGS, UA-CVAE(M) and UA-CVAE(C) achieved a 11\% and 17\% gain. The difference in gains could be attributed to the informative nature and shorter length of the dialogue utterances in ConvAI2, which would result in more positive NLI predictions (`entailment') when computing the UE-score. Also, we notice that UA-CVAE(C) has a slight edge over UA-CVAE(M) in terms of coherence. In addition to attaining higher UE-scores, the UA-CVAE(C) model achieved a larger percentage of wins on both corpora. This shows that a 1D convolution layer allows for a slightly more effective combination of the variance, or uncertainty, vector and latent variable.

\section{Conclusion}
We have proposed two UA-CVAE frameworks: UA-CVAE(M) and UA-CVAE(C), which improved coherence on both ConvAI2 (60\% and 121\%) and EMPATHETICDIALOGS (11\% and 17\%). Additionally, we introduced the UE-score, which correlated positively with human judgement in terms of contextual coherence. Future work could involve exploring the possibility of enhancing the dialogue quality by leveraging the epistemic or homoscedastic aleatoric uncertainty.

\vfill
\bibliographystyle{IEEEbib}
\bibliography{strings,refs}

\end{document}